\documentclass[11pt]{article}

\usepackage[preprint]{acl}

\usepackage{times}
\usepackage{latexsym}

\usepackage{tabularx}

\usepackage[T1]{fontenc}

\usepackage[utf8]{inputenc}

\usepackage{microtype}

\usepackage{inconsolata}

\usepackage{graphicx}

\usepackage{amssymb} 

\usepackage[most]{tcolorbox}
\usepackage[dvipsnames]{xcolor}

\usepackage{multirow}
\usepackage{multicol}
\usepackage{pifont}
\usepackage{latexsym}
\newcommand{\cmark}{\textcolor{teal}{\ding{51}}}%
\newcommand{\xmark}{\textcolor{purple}{\ding{55}}}%

\newcommand{\Stwo}[0]{\textcolor{OliveGreen}{\textbf{Step 2 }}}
\newcommand{\Sone}[0]{\textcolor{NavyBlue}{\textbf{Step 1 }}}
\newcommand{\Sthree}[0]{\textcolor{Orange}{\textbf{Step 3 }}}

\usepackage{url}
\usepackage{color}

\title{A Simple Method to Enhance Pre-trained Language Models with Speech Tokens for Classification}

  \author{Nicol\'{a}s Calbucura \\
  Universidad de Chile, DCC \\
  Santiago, Chile \\
  \texttt{\small{nicolas.calbucura@ug.uchile.cl}} \\\And
  Jos\'{e} Guillen \\
  UTFSM \\
  Valparaiso, Chile \\
  \texttt{\small{jose.guillen@cenia.cl}} \\\And
  Valentin Barriere \\
  Universidad de Chile, DCC, CENIA \\
  Santiago, Chile \\
  \texttt{\small{vbarriere@dcc.uchile.cl}} \\}

\begin{document}
\maketitle
\begin{abstract}
This paper presents a simple method that allows to easily enhance textual pre-trained large language models with speech information, when fine-tuned for a specific classification task. 
A classical issue with the fusion of many embeddings from audio with text is the large length of the audio sequence compared to the text one. 
Our method benefits from an existing speech tokenizer trained for Audio Speech Recognition that output long sequences of tokens from a large vocabulary, making it difficult to integrate it at low cost in a large language model. 
By applying a simple lasso-based feature selection on multimodal Bag-of-Words representation, we retain only the most important audio tokens for the task, and adapt the language model to them with a self-supervised language modeling objective, before fine-tuning it on the downstream task. We show this method improves the performances compared to an unimodal model, larger SpeechLMs, and audio integration via learned representations.  
%
We demonstrate its effectiveness on Argumentative Fallacy Detection and Classification tasks where audio was previously believed counterproductive, and affective computing tasks on a widely-used dataset. 
We also provide an in-depth analysis of the method, showing that even a random audio token selection helps enhancing the unimodal model. Our code is available 
\href{https://github.com/salocinc/EMNLP26SpeechTokLLM/}{online}.
%
\end{abstract}

\begin{figure*}[t]  
\centering
\includegraphics[width=1.\textwidth]{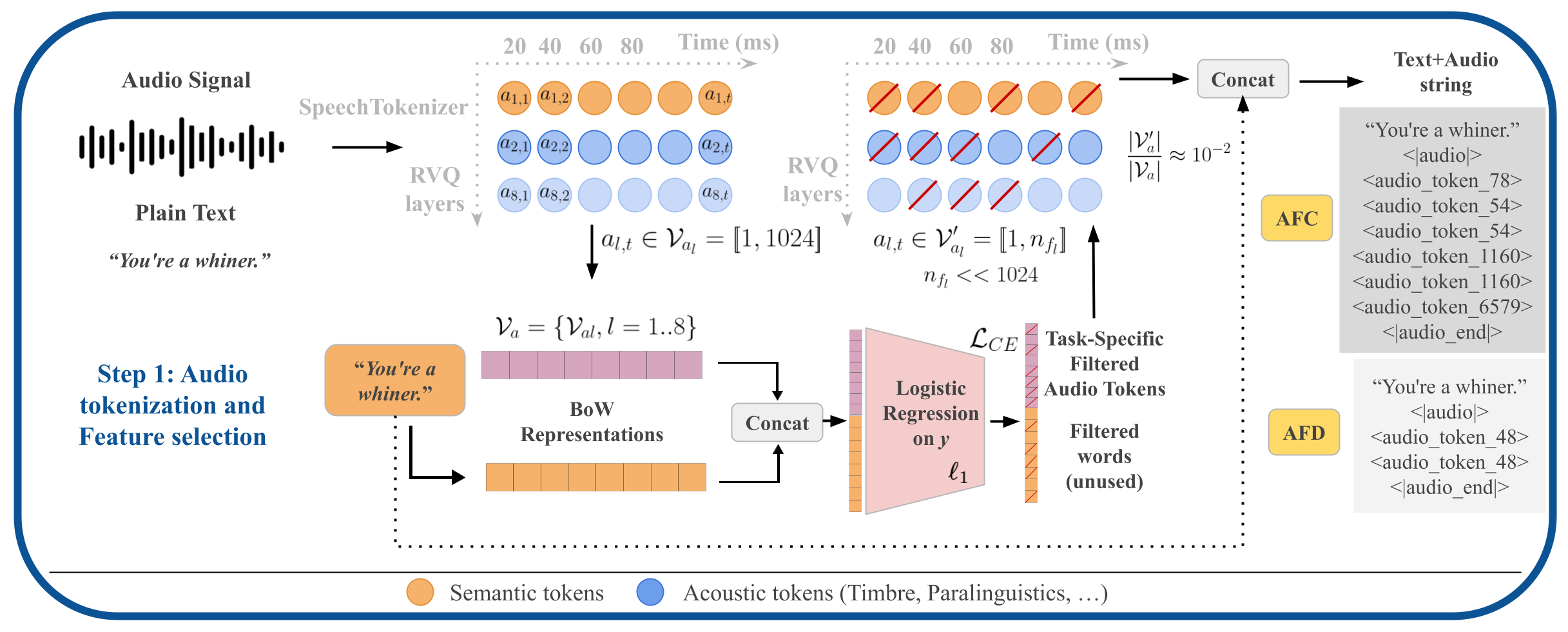}  
\caption{
\Sone of our method: audio token selection via $\ell_1$-regularized logistic regression on Bag-of-Words representations. Given an audio signal, SpeechTokenizer produces tokens $a_{l,t}$ at each layer $l$ and time step $t$ from a vocabulary $\mathcal{V}_a$ composed of $\mathcal{V}_{a{_l}}$ of size $|\mathcal{V}_{a{_l}}| = 1{,}024$. The $\ell_1$ selection retains $n_{f_l}$ features per layer, yielding a reduced vocabulary $\mathcal{V}'_a \subset \mathcal{V}_a$ with $|\mathcal{V}'_a| \approx 0.01 \cdot |\mathcal{V}_a|$.}  \vspace*{-.3cm}



\label{fig:feature_selection}
\end{figure*}

\section{Introduction and Related Work}



Human communication is composed of several modalities, which play different roles on the information transmission and can interact with each other \cite{Vinciarelli2009a}. \citet{Frohlich2019} state that human communication and non-human primate communication is inherently multimodal, which makes necessary to put efforts on the multimodal analysis of natural language.

In recent years, multiple fields of Natural Language Processing have seen advances on the multimodality path. For example, vision-language models such as CLIP \cite{Radford2021} and Flamingo \cite{Alayrac2022} have demonstrated strong performance on image-text understanding tasks. Furthermore, Argument Mining field has also taken steps towards the multimodality. Efforts such as \citet{Lippi2016,Mestre2021,Mancini2022,Mestre2023} 
have paved the way to explore the benefits of modality representations and fusion methods.

Methods integrating the audio such as SpeechVerse, Qwen2-Audio or Audio-Flamingo \cite{Das2024,Chu2024,Kong2024} are based on several pre-training stages, composed of ASR followed by instruction tuning on several tasks. If implemented directly on a target task with a small dataset, they fail \cite{Thimonier2025}. 
%
%

Another important obstacle to integrating speech tokens into text-based LLMs is the \emph{sequence length mismatch}: speech and audio representations are generated at significantly higher temporal frequencies than text (typically 50\,Hz versus approximately 2\,Hz for natural language \cite{Das2024,Defossez2024,Mancini2024a}). This disparity leads to audio token sequences that are an order of magnitude longer than their textual counterparts (50\,Hz across multiple layers vs.\ $\sim$2\,Hz for text), introducing severe attention sparsity and computational overhead during LLM processing \cite{Liu2025,Mancini2024a}. 

The method presented in this work uses Speech Tokenization in order to include acoustic information into an LLM, which is defined as the process of converting continuous audio signals into discrete representations (tokens) that capture semantic and/or acoustic information, allowing them to be processed by models originally designed for text or other discrete inputs. 
The tokens fall into two main categories \cite{Borsos2023}: semantic tokens and acoustic tokens, which are typically obtained separately: for instance, HuBERT \cite{Hsu2021} produces semantic tokens, while EnCodec \cite{Defossez2022} generates acoustic tokens. 
Instead of compressing the sequence, our method identifies the \textit{specific acoustic tokens that carry the most information for the downstream task.}

In this paper, we focus on the Multimodal Argument Mining field, specifically on the Argumentative Fallacy Classification (AFC) and Argumentative Fallacy Detection (AFD) tasks using datasets 
from \citet{Mancini2024a}. As baseline, we consider the results presented in 
\citet{Mancini2025}, which showed a strong text dominance over audio and did not find benefits from multimodal fusion.

\paragraph{Motivation and Contribution}
%
Adapting LLMs to speech suffers from audio-text length mismatches, requiring massive pre-training data. 
While inference-time pruning operates dynamically on instance-level attention patterns, we propose a complementary, \emph{task-level} approach: we apply $\ell_1$-regularized logistic regression on multimodal Bag-of-Words\footnote{composed of both textual and audio tokens} representations to select only the audio token types that carry discriminative information for the downstream task, reducing the vocabulary to less than 1\% of its original size \emph{prior to fine-tuning}. The selected tokens are then adapted to the LLM through a lightweight self-supervised pre-training step before task fine-tuning. Unlike large-scale SpeechLLM pipelines \cite{Das2024,Chu2024,Kong2024}, our method requires no multi-stage pre-training and can be applied to any pre-trained LLM at minimal computational cost.

\section{Method}

\begin{figure*}[t]  
\centering
\includegraphics[width=1\textwidth]{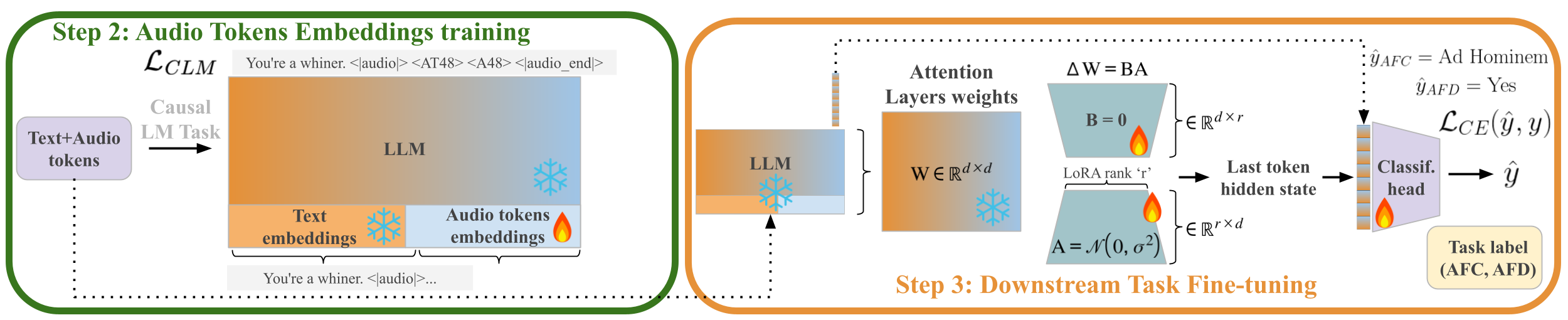}  
\caption{The \Stwo and \Sthree consist in the pretraining the audio tokens embeddings followed by the fine-tuning of the multimodal LLM on the downstream task.}  \vspace*{-.3cm}
\label{fig:pretraining}
\end{figure*}

Our method fuses speech-related information with text by means of a pretrained LLM and audio tokens (AT) from a trained speech tokenizer (section \ref{subsec:tokenizer}). First,  we process to a task-specific feature selection of the tokens in \Sone to reduce the length of the sequence and size of the vocabulary (Section \ref{subsec:selection} and Figure \ref{fig:feature_selection}). Then, the tokens are adapted to the LLM in \Stwo by continuing the pretraining (Section \ref{subsec:pretraining}).   
Finally, the model is fine-tuned on the downstream task in \Sthree using a classification head (Section \ref{subsec:downstream} and Figure \ref{fig:pretraining}). 


\subsection{Speech Tokenization} \label{subsec:tokenizer}

We process to the tokenization using \textit{SpeechTokenizer} \cite{Zhang2024}, which is an acoustic-semantic unified Encoder-Decoder architecture based on Residual Vector Quantization (RVQ; \citet{Gray1985, Vasuki2006}). Its encoder generates a hierarchical latent representation: the first layer or quantizer 
captures semantic content, while the following residual layers 
capture paralinguistic content. 
%

To reduce the redundancy between the modalities, we use audio tokens obtained with \textit{SpeechTokenizer} considering all of its 8 layers or only the last 7 layers. Since the first layer (RVQ-1) encodes semantic content that is already captured by the text modality, excluding it aims to integrate only paralinguistic information from the speech (such as prosody, rhythm, and voice quality) while leaving the semantic content to the text. We empirically validate this design choice in Section~\ref{sec:analysis} (Table~\ref{tab:ablationtokens}).


The output of the tokenizer is 2-dimensional. On the first dimension, it has a vocabulary $\mathcal{V}_{al}$ of size 1024 at each layer level $l$, which gives a total of 8,196 different tokens in the initial vocabulary $\mathcal{V}_a$.\footnote{It would be technically $1024^8$ if taking all combinations.} The second dimension depends on the length and content of the audio sequence: one token per layer/quantizer is produced every 20 milliseconds. 

\subsection{Token Selection} \label{subsec:selection}

A classical problem with the multimodal fusion of audio with text is the sampling inconsistency as the audio sampling frequency is generally way higher than words \cite{Zadeh2018b,Das2024}.\footnote{In \textit{SpeechTokenizer} case, an audio token every 20ms} This situation can create an overflow of audio content compared to textual content \cite{Mancini2024a}, with an amount of audio tokens obtained from an audio signal significantly bigger than the amount of text tokens. 
%
This disproportion may cause semantic and attention sparsity \cite{Liu2025}, negatively affecting the model performance due to audio features dominating the early stages of training, leading to underwhelming performance.


For this reason, an audio token selection pipeline is implemented as a final goal to reduce the amount of AT that will be concatenated with samples' plain text. The process is showed in Figure~\ref{fig:feature_selection} and consists of: \textit{(i)} tokenizing the audio signal; \textit{(ii)} representing both text and audio using Bag-of-Word representations. For audio, each layer contributes a 1024-dimensional histogram, yielding a $1024 \times L$-dimensional vector per sample (where $L$ is the number of layers), which is concatenated with the textual BoW; \textit{(iii)} selecting the most relevant features by training a $\ell_1$-regularized logistic regression to predict the task label. The $\ell_1$ penalty drives the coefficients of irrelevant audio dimensions to zero, effectively selecting only the tokens that carry task-discriminative information beyond what the text already provides; \textit{(iv)} filter the audio tokens to keep only the selected ones. 

The BoW representation is used exclusively for this selection step, not for the downstream model: it provides a simple, order-agnostic summary sufficient for identifying \emph{which} token types are informative, while the temporal structure of the surviving tokens is preserved and modeled by the LLM during fine-tuning.
The size of the final vocabulary $\mathcal{V}'_a$ is less then 1\% of $\mathcal{V}_a$, and the selection is task-specific, performed on the training split only.

\subsection{Audio Tokens embeddings Pre-training} \label{subsec:pretraining}


The addition of the selected audio tokens to the LLM means extending its vocabulary with $|\mathcal{V}'_a|$ new entries. Each new audio token is a one-hot vector mapped to a randomly initialized embedding of the same dimensionality as the LLM's existing token embeddings. No additional projection layers are introduced. We then adapt these embeddings using a causal language modeling cross-entropy loss $\mathcal{L}_{CLM}$ (Equation~\ref{eq:CLM}). During this pre-training phase, all original LLM parameters remain frozen; only the new audio token embeddings are updated, ensuring that the model's pre-trained linguistic representations are preserved (Figure~\ref{fig:pretraining}, \Stwo).




\begin{equation}
    \mathcal{L}_{CLM} = \sum_t p(a_t|a_{<t}, w_1...w_n)
    \label{eq:CLM}
\end{equation}

\subsection{Downstream Task Fine-tuning} \label{subsec:downstream}

\begin{table*}
\centering
\resizebox{.9\textwidth}{!}{
\begin{tabular}{llcccc}
\cline{1-6}
\textbf{Task} & \textbf{Method} & \textbf{Bagging} & \textbf{Text} & \textbf{Audio} & \textbf{Text + Audio} \\
\hline \hline
\multirow{10}{*}{AFC} & RoBERTa + Whisper$^\dagger$ \cite{tahir-etal-2025-prompt} & \multirow{3}{*}{\xmark} & 0.486 & 0.159 & 0.461 \\
 & RoBERTa + HuBERT$^\dagger$ \cite{pittiglio-2025-leveraging} & & 0.444 & \textbf{0.356} & 0.440 \\
 & (MM-)RoBERTa + WavLM$^\dagger$ &  & 0.393 & 0.064 & 0.382 \\ \cline{2-6}
 & Qwen2-Audio-7B (ICL) \cite{Chu2024} & \multirow{4}{*}{\xmark} & -- & -- & 0.170 \\
 & Qwen2-Audio-7B (SFT) \cite{Chu2024} & & -- & -- & 0.507 \\
 & Llama3-3B+BYOL-A (SFT) & & 0.523 & -- & 0.504 \\
 & Our method & & -- & -- & 0.548 \\ \cline{2-6}
 & Qwen2-Audio-7B (SFT) \cite{Chu2024} & \multirow{3}{*}{\cmark} & -- & -- & 0.515 \\
 & Llama3-3B+BYOL-A (SFT) & & \textbf{0.537} & -- & 0.547 \\
 & Our method & & -- & -- & \textbf{0.592} \\
\hline \hline
\multirow{6}{*}{AFD} & RoBERTa + wav2vec2.0$^\dagger$ \cite{cantin-larumbe-chust-vendrell-2025-argumentative} & \multirow{2}{*}{\xmark} & 0.220 & 0.169 & 0.193 \\
 & (MM-)RoBERTa + WavLM$^\dagger$ & & 0.277 & 0.000 & 0.285 \\ \cline{2-6}
 & Llama3-3B+BYOL-A (SFT) & \multirow{2}{*}{\xmark} &0.283 & -- & 0.285 \\
 & Our method & &  --  & -- & 0.307 \\ \cline{2-6}
 & Llama3-3B+BYOL-A (SFT) & \multirow{2}{*}{\cmark} & \textbf{0.297} & -- & 0.301 \\
 & Our method  &  & -- & -- & \textbf{0.317}\\
\hline
\end{tabular}
}
\caption{Results for AFD and AFC tasks in terms of Macro-F1. 
Best baselines or SoTA are reported. $^\dagger$ indicates the results come from \citet{Mancini2025}. ICL and SFT stand for In-Context-Learning and Supervised Fine-Tuning.}  
\label{tab:results}
\end{table*}

Once the new Audio Tokens embeddings are trained, the LLM is fine-tuned using LoRA and a classification head consisting of a single linear layer. This head maps the final hidden representation of the last token to a logits vector which has the quantity of target classes in the downstream classification task (e.g., 2 for AFD: "No Fallacy" and "Fallacy"). Only the LoRA adapter parameters and the classification head are updated during fine-tuning, while the rest of the model remains frozen to preserve the pretrained linguistic and acoustic knowledge encoded in the base LLM.

\section{Experiment and Results}

\subsection{Experimental Protocol}

We validate our method on speech classification tasks: fallacy detection (AFD) and classification (AFC), sentiment analysis (SENT), and emotion classification (EMO). 
%

We use the \textbf{MM-USED} dataset \cite{Mancini2022} to train the Audio Token embeddings, the \textbf{MM-USED-Fallacy} dataset \cite{Mancini2024,Mancini2025} for fallacy tasks, and the \textbf{CMU-MOSEI} dataset \cite{Zadeh2018d} for SENT and EMO tasks.
Results on fallacy tasks are reported on the official test set of the MM-ArgFallacy2025 shared task \cite{Mancini2025}, which uses MM-USED-Fallacy for training and validation with the same splits as the shared task challenge.
For the AT selection, $\ell_1$ penalization ensures token length matching between audio and text. We integrate BYOL-A representations by projecting them onto the LLM embedding space. 
Following \cite{Mancini2025}, we use macro-avg F1-score for AFC and F1-score of the positive class for AFD. 
Results are averaged over 5 seeds with probability-based bagging. Dataset details 
are provided in Appendix~\ref{app:datasets}, and baseline descriptions in Appendix~\ref{app:baselines}.

\subsection{Results} 

The results are visible in Tables \ref{tab:results_mosei} and \ref{tab:results}. It is possible to see that our method outperforms known state-of-the-art on this dataset, but also other systems based on models 2.5 larger such as \texttt{Qwen2-Audio-7B(-Instructed)}. 

\begin{table}
\centering
\resizebox{.5\textwidth}{!}{
\begin{tabular}{llcc}
\textbf{Task} & \textbf{Method} & \textbf{Text} &  \textbf{Text + Audio} \\
\hline \hline
  \multirow{3}{*}{SENT} & Qwen2-Audio-7B (SFT)  &  -- & 0.779\\
  & Llama3-3B+BYOL-A (SFT)& 0.764 & 0.771 \\
 & Our method  &  -- & \textbf{0.792}\\
 \hline \hline
  \multirow{3}{*}{EMO} & Qwen2-Audio-7B (SFT)  &  -- & 0.734\\
  & Llama3-3B+BYOL-A (SFT)& 0.726 & 0.732 \\
 & Our method  &  -- & \textbf{0.742}\\
\hline
\end{tabular}
}
\caption{Results on CMU-MOSEI Sentiment analysis and Emotion classification tasks in terms of Macro-F1. 
}  
\label{tab:results_mosei}
\end{table}

\paragraph{Sentiment and Emotion Classification}
Table \ref{tab:results_mosei} shows the results on CMU-MOSEI, comparing an unimodel LLaMA-3B with our model. In both tasks, our method helps improving the results (75.8 to 79.2 and 71.3 to 74.2).

\paragraph{Argumentative Fallacy Classification}
Analyzing the results of the uni- and multimodal models on the AFC task, we observe that LLaMA-3B achieves highly competitive performance when fine-tuned on the training set, surpassing the previous state-of-the-art (53.7 vs. 48.6). Incorporating audio tokens further boosts performance (from 52.3 to 54.8). 
Finally, 
bagging further improves the multimodal model more than the unimodal one 
(+4.4\% vs. +1.4\%). The best results are obtained with the multimodal configuration, exceeding the prior state-of-the-art by more than 10 points.




\paragraph{Argumentative Fallacy Detection}
Even though binary, AFD is more challenging than AFC due to class unbalance. The initial performance between a RoBERTa and a Llama 30 times bigger questions the usefulness of such a method. However, the gap is increasing when adding the audio modality, which does not manage well the MM-RoBERTa.

\section{Analysis}\label{sec:analysis} 


\paragraph{Speech Tokens Selection Methods} We ran experiments with respect to the different methods used to select the speech tokens. Results are shown in Table \ref{tab:ablationtokens} where  
4 different configurations were compared to investigate the impact of the audio token type (acoustic or semantic), pre-training or the feature selection. 
We picked the AT from the 8 layers, using both semantic and acoustic tokens, or only the acoustic tokens (last 7 layers), comparing lasso-based vs random selection. 

\begin{table}[ht]
    \centering
    \resizebox{.4\textwidth}{!}{
    \begin{tabular}{c|c|c|c}
        Audio PT & Filtering Semantic & $\ell_1$ & F1 \\ \hline \hline
        $\varnothing$ &  $\varnothing$ & $\varnothing$ & 52.3 
        \\ \hline
        \xmark &  \cmark  & \xmark & 48.7 
        \\
        \xmark &  \cmark  & \cmark & 52.7
        \\ \hline
        \cmark &  \xmark  & \xmark & 53.3 
        \\
        \cmark &  \xmark  & \cmark & 51.8
        \\
        \cmark &  \cmark & \xmark & 53.9 
        \\
        \cmark &  \cmark & \cmark & \textbf{54.8} 
        \\
        
    \end{tabular}
    }
    \caption{Results on the AFC task using different AT selection with respect to acoustic or semantic tokens, and $\ell_1$-based selection versus random selection. 
    }  
    \label{tab:ablationtokens}
\end{table}

It is first notable that the use of audio tokens always improve the results, even by selecting them randomly, supposing that complementary information is always contained in the audio tokens. 
Finally, the model trained with acoustic-only tokens reach higher performances than  the ones using also semantic tokens. Which is aligned with the idea that the LLM is better at processing the text that contains the semantic information.


\paragraph{Pre-training and $\ell_1$ Effect}
The model trained without pre-training benefits from the lasso-based AT filtering, as without it, it reaches low F1 of .487 with a very standard deviation 5 times higher than other configurations. This implies that if PT is not necessary, it still needs AT selection. 

\paragraph{Audio Vocabulary Size} 

The average length of a textual token sequence is 31.52 while the average length of a speech token sequence is 35.52, which is around the same order. The number of tokens selected is 73 for the 7-layers configuration and 80 for the full 8-layers configuration. Which is both case is $<$ 1\% of the initial vocabulary size.

\paragraph{Token Redundancy}
Our findings align with recent observations on token redundancy in audio-language architectures. \citet{Gibier2025} demonstrated that encoder attention in audio-language models concentrates on a narrow subset of informative frames, and that retaining as few as 25\% of audio tokens preserves performance on captioning and question-answering benchmarks. Our results corroborate this from a complementary angle: filtering the audio vocabulary down to less than 1\% via $\ell_1$-regularized selection not only maintains classification performance but consistently improves it over unimodal baselines (Tables~\ref{tab:results_mosei} and~\ref{tab:results}). This suggests that the vast majority of tokens produced by fine-grained speech tokenizers carry information that can be safely discarded for many tasks, and that targeted selection at the vocabulary level is an effective strategy for integrating speech into text-based LLMs.




\paragraph{Fine-Grained Examples Analysis} 
Analysing the discrepancy between the predictions of the unimodal textual model with the predictions of the speech enhanced one, we saw that the vanilla model struggles to classify \textit{Appeal to Emotion} when the sentence has presence of "\textit{He}" or "\textit{She}", resulting on a \textit{Ad Hominem} prediction. We also detected the presence of audio tokens on these cases that were more likely to appear on \textit{Appeal to Emotion} than other fallacy classes, which could indicate those tokens are class-representative. 
Analysing examples in MOSEI, we found the multimodal model took advantage of acoustic cues when text only was not enough to disambiguate between the classes:
examples with neutral text with happy or sad voice. 




\section{Conclusion} 

We propose a new method to integrate audio cues inside a pre-trained transformer, in order to enhance it with specific audio tokens. We handle the high frequency of audio tokens by selecting task-relevant ones using a lasso-based logistic regression. We demonstrate the effectiveness of this method on four classification tasks across two domains where audio was previously believed not useful, achieving state-of-the-art results against strong baselines including a 2.5$\times$ larger multimodal model. While our evaluation is focused on classification settings with a single LLM backbone, the consistent improvements suggest that task-specific audio token selection is a promising direction for lightweight multimodal adaptation of pre-trained LLMs.


\section*{Limitations}


While the proposed method demonstrates that incorporating selected speech tokens can effectively enhance textual large language models on classification tasks, several limitations remain.

\paragraph{Scope of Empirical Validation}
Our evaluation covers four classification tasks across two datasets (MM-USED-Fallacy and CMU-MOSEI). While these span different domains (argument mining and affective computing) and demonstrate consistent improvements, we acknowledge that broader validation on additional datasets, languages, and task types would strengthen the generalizability of our findings. Similarly, our method has been evaluated with a single LLM backbone (LLaMA-3.2-3B) and a single speech tokenizer (SpeechTokenizer); experiments with other architectures would further validate its general applicability. We frame our contribution as a lightweight strategy for adapting text-based LLMs with selected audio tokens, rather than a fully general multimodal token-selection framework.
Extending the approach to language generation tasks, such as those explored in \cite{Kim2025} for gestures and facial expressions, could provide further insights into its generalization ability.


Another limitation is about how to integrate the audio tokens: we fine-tune the audio embeddings jointly with the downstream task. Evaluating the approach on downstream tasks without fine-tuning these embeddings would help assess their intrinsic generalization capacity. 
Similarly, we fine-tune the only the embedding layers using causal language modeling; future work could investigate other less efficient strategies, such as updating all parameters. 
Finally, it would be interesting to experiment with a lower lasso regularization to get a larger number of speech tokens and empirically find the optimal size.

Beyond model-level analysis, it would be valuable to conduct a correlational study between the learned speech tokens and expert acoustic features such as pitch, intensity, spectral shape or features of the glottal flow \cite{Degottex2014}. This could improve interpretability and guide future tokenization strategies. 
In addition, a deeper attention flow analysis could clarify how the model integrates acoustic information across layers.

It would be useful to augment the LLM by integrating information from political debate forums or public consultations with respect to political issues \cite{Barriere2023,Barriere2022}. 

Finally, our experimental setup could be broadened by including other SpeechLLM baselines (e.g., Audio-Flamingo2-3B or EmoSLLM \cite{Kong2024,Thimonier2025}) to compare, but also other textual LLMs, though this is beyond the scope of this paper. 

\section*{Acknowledgments}
The authors thank the reviewers for the various comments that helped to improve the manuscript. 
This work was partially financed by the ANID fondecyt grant 11251024 "Multimodal Argumentation Mining in Groups Assisted by an Embodied Conversational Agent" and 
by National Center for Artificial Intelligence CENIA FB210017, Basal ANID.

\bibliography{custom,humour_standup,JRC}

\begin{thebibliography}{38}
\providecommand{\natexlab}[1]{#1}

\bibitem[{Abadi et~al.(2016)Abadi, Barham, Chen, Chen, Davis, Dean, Devin,
  Ghemawat, Irving, Isard, Kudlur, Levenberg, Monga, Moore, Murray, Steiner,
  Tucker, Vasudevan, Warden, Wicke, Yu, and Zheng}]{Abadi2016}
Martín Abadi, Paul Barham, Jianmin Chen, Zhifeng Chen, Andy Davis, Jeffrey
  Dean, Matthieu Devin, Sanjay Ghemawat, Geoffrey Irving, Michael Isard,
  Manjunath Kudlur, Josh Levenberg, Rajat Monga, Sherry Moore, Derek~G. Murray,
  Benoit Steiner, Paul Tucker, Vijay Vasudevan, Pete Warden, and 3 others.
  2016.
\newblock {TensorFlow: A system for large-scale machine learning}.
\newblock \emph{Proceedings of the 12th USENIX Symposium on Operating Systems
  Design and Implementation, OSDI 2016}, pages 265--283.

\bibitem[{Alayrac et~al.(2022)Alayrac, Donahue, Luc, Miech, Barr, Hasson, Lenc,
  Mensch, Millican, Reynolds, Ring, Rutherford, Han, Gong, Samangooei,
  Monteiro, Menick, Borgeaud, Brock, Nematzadeh, Sharifzadeh, Binkowski,
  Barreira, Vinyals, Zisserman, and Simonyan}]{Alayrac2022}
Jean~Baptiste Alayrac, Jeff Donahue, Pauline Luc, Antoine Miech, Iain Barr,
  Yana Hasson, Karel Lenc, Arthur Mensch, Katie Millican, Malcolm Reynolds,
  Roman Ring, Eliza Rutherford, Serkan Cabi~Tengda Han, Zhitao Gong, Sina
  Samangooei, Marianne Monteiro, Jacob Menick, Sebastian Borgeaud, Andrew
  Brock, and 7 others. 2022.
\newblock {Flamingo: a Visual Language Model for Few-Shot Learning}.
\newblock In \emph{Advances in Neural Information Processing Systems},
  volume~35.

\bibitem[{Barriere and Balahur(2023)}]{Barriere2023}
Valentin Barriere and Alexandra Balahur. 2023.
\newblock \href {https://www.mdpi.com/2227-7390/11/9/2161} {{Multilingual
  Multi-target Stance Recognition in Online Public Consultations}}.
\newblock \emph{MDPI Mathematics -- Special issue on Human Language
  Technollogy}, 11(9):2161.

\bibitem[{Barriere and Jacquet(2022)}]{Barriere2022}
Valentin Barriere and Guillaume Jacquet. 2022.
\newblock {CoFE : A New Dataset of Intra-Multilingual Multi-target Stance
  Classification from an Online European Participatory Democracy Platform}.
\newblock \emph{AACL-IJCNLP}.

\bibitem[{Borsos et~al.(2023)Borsos, Marinier, Vincent, Kharitonov, Pietquin,
  Sharifi, Roblek, Teboul, Grangier, Tagliasacchi, and Zeghidour}]{Borsos2023}
Zalán Borsos, Raphaël Marinier, Damien Vincent, Eugene Kharitonov, Olivier
  Pietquin, Matt Sharifi, Dominik Roblek, Olivier Teboul, David Grangier, Marco
  Tagliasacchi, and Neil Zeghidour. 2023.
\newblock \href {https://arxiv.org/abs/2209.03143} {Audiolm: a language
  modeling approach to audio generation}.
\newblock \emph{Preprint}, arXiv:2209.03143.

\bibitem[{Cant{\'{i}}n and
  Chust(2025)}]{cantin-larumbe-chust-vendrell-2025-argumentative}
Eva Cant{\'{i}}n and Adriana Chust. 2025.
\newblock \href {https://doi.org/10.18653/v1/2025.argmining-1.36}
  {{Argumentative Fallacy Detection in Political Debates}}.
\newblock In \emph{Proceedings of the 12th Argument mining Workshop}, pages
  369--373, Vienna, Austria. Association for Computational Linguistics.

\bibitem[{Chu et~al.(2024)Chu, Xu, Yang, Wei, Wei, Guo, Leng, Lv, He, Lin,
  Zhou, and Zhou}]{Chu2024}
Yunfei Chu, Jin Xu, Qian Yang, Haojie Wei, Xipin Wei, Zhifang Guo, Yichong
  Leng, Yuanjun Lv, Jinzheng He, Junyang Lin, Chang Zhou, and Jingren Zhou.
  2024.
\newblock \href {http://arxiv.org/abs/2407.10759} {{Qwen2-Audio Technical
  Report}}.
\newblock pages 1--16.

\bibitem[{Das et~al.(2024)Das, Dingliwal, Ronanki, Paturi, Huang, Mathur, Yuan,
  Bekal, Niu, Jayanthi, Li, Mundnich, Sunkara, Bodapati, Srinivasan, Han, and
  Kirchhoff}]{Das2024}
Nilaksh Das, Saket Dingliwal, Srikanth Ronanki, Rohit Paturi, Zhaocheng Huang,
  Prashant Mathur, Jie Yuan, Dhanush Bekal, Xing Niu, Sai~Muralidhar Jayanthi,
  Xilai Li, Karel Mundnich, Monica Sunkara, Sravan Bodapati, Sundararajan
  Srinivasan, Kyu~J Han, and Katrin Kirchhoff. 2024.
\newblock \href {http://arxiv.org/abs/2405.08295} {{SpeechVerse: A Large-scale
  Generalizable Audio Language Model}}.

\bibitem[{D{\'{e}}fossez et~al.(2024)D{\'{e}}fossez, Mazar{\'{e}}, Orsini,
  Royer, P{\'{e}}rez, J{\'{e}}gou, Grave, and Zeghidour}]{Defossez2024}
Alexandre D{\'{e}}fossez, Laurent Mazar{\'{e}}, Manu Orsini, Amélie Royer,
  Patrick P{\'{e}}rez, Hervé J{\'{e}}gou, Edouard Grave, and Neil Zeghidour.
  2024.
\newblock \href {http://arxiv.org/abs/2410.00037} {{Moshi: a speech-text
  foundation model for real-time dialogue}}.
\newblock pages 1--67.

\bibitem[{Degottex et~al.(2014)Degottex, Kane, Drugman, Raitio, and
  Scherer}]{Degottex2014}
Gilles Degottex, John Kane, Thomas Drugman, Tuomo Raitio, and Stefan Scherer.
  2014.
\newblock \href {https://doi.org/10.1109/ICASSP.2014.6853739} {{COVAREP - A
  collaborative voice analysis repository for speech technologies}}.
\newblock In \emph{ICASSP, IEEE International Conference on Acoustics, Speech
  and Signal Processing - Proceedings}, pages 960--964.

\bibitem[{Défossez et~al.(2022)Défossez, Copet, Synnaeve, and
  Adi}]{Defossez2022}
Alexandre Défossez, Jade Copet, Gabriel Synnaeve, and Yossi Adi. 2022.
\newblock \href {https://arxiv.org/abs/2210.13438} {High fidelity neural audio
  compression}.
\newblock \emph{Preprint}, arXiv:2210.13438.

\bibitem[{Fr{\"{o}}hlich et~al.(2019)Fr{\"{o}}hlich, Sievers, Townsend, Gruber,
  and van Schaik}]{Frohlich2019}
Marlen Fr{\"{o}}hlich, Christine Sievers, Simon~W. Townsend, Thibaud Gruber,
  and Carel~P. van Schaik. 2019.
\newblock \href {https://doi.org/10.1111/brv.12535} {{Multimodal communication
  and language origins: integrating gestures and vocalizations}}.
\newblock \emph{Biological Reviews}, 94(5):1809--1829.

\bibitem[{Gibier et~al.(2025)Gibier, Duroselle, Serrano, Boeffard, and
  Bonastre}]{Gibier2025}
Marcel Gibier, Rapha{\"e}l Duroselle, Pierre Serrano, Olivier Boeffard, and
  Jean-Fran{\c{c}}ois Bonastre. 2025.
\newblock \href {https://arxiv.org/abs/2511.14293} {Segmentwise pruning in
  audio-language models}.
\newblock In \emph{Proceedings of ICASSP 2026 -- IEEE International Conference
  on Acoustics, Speech and Signal Processing}.

\bibitem[{Goffredo et~al.(2022)Goffredo, Haddadan, Vorakitphan, Cabrio, and
  Villata}]{Goffredo2022}
Pierpaolo Goffredo, Shohreh Haddadan, Vorakit Vorakitphan, Elena Cabrio, and
  Serena Villata. 2022.
\newblock \href {https://doi.org/10.24963/ijcai.2022/575} {{Fallacious Argument
  Classification in Political Debates}}.
\newblock \emph{IJCAI International Joint Conference on Artificial
  Intelligence}, pages 4143--4149.

\bibitem[{Gray(1984)}]{Gray1985}
R.~Gray. 1984.
\newblock \href {https://doi.org/10.1109/MASSP.1984.1162229} {Vector
  quantization}.
\newblock \emph{IEEE ASSP Magazine}, 1(2):4--29.

\bibitem[{Haddadan et~al.(2019)Haddadan, Cabrio, and Villata}]{Haddadan2019}
Shohreh Haddadan, Elena Cabrio, and Serena Villata. 2019.
\newblock \href {https://doi.org/10.18653/v1/p19-1463} {{Yes, we can! Mining
  arguments in 50 years of US presidential campaign debates}}.
\newblock \emph{ACL 2019 - 57th Annual Meeting of the Association for
  Computational Linguistics, Proceedings of the Conference}, pages 4684--4690.

\bibitem[{Hsu et~al.(2021)Hsu, Bolte, Tsai, Lakhotia, Salakhutdinov, and
  Mohamed}]{Hsu2021}
Wei~Ning Hsu, Benjamin Bolte, Yao Hung~Hubert Tsai, Kushal Lakhotia, Ruslan
  Salakhutdinov, and Abdelrahman Mohamed. 2021.
\newblock \href {https://doi.org/10.1109/TASLP.2021.3122291} {{HuBERT:
  Self-Supervised Speech Representation Learning by Masked Prediction of Hidden
  Units}}.
\newblock \emph{IEEE/ACM Transactions on Audio Speech and Language Processing},
  29(Cv):3451--3460.

\bibitem[{Kim et~al.(2025)Kim, Chung, Kim, Lee, Lee, Kim, Yang, and
  Yu}]{Kim2025}
Youngmin Kim, Jiwan Chung, Jisoo Kim, Sunghyun Lee, Sangkyu Lee, Junhyeok Kim,
  Cheoljong Yang, and Youngjae Yu. 2025.
\newblock \href {https://doi.org/10.18653/v1/2025.acl-long.112} {{Speaking
  Beyond Language: A Large-Scale Multimodal Dataset for Learning Nonverbal Cues
  from Video-Grounded Dialogues}}.
\newblock pages 2247--2265.

\bibitem[{Kong et~al.(2024)Kong, Goel, Badlani, Ping, Valle, and
  Catanzaro}]{Kong2024}
Zhifeng Kong, Arushi Goel, Rohan Badlani, Wei Ping, Rafael Valle, and Bryan
  Catanzaro. 2024.
\newblock {Audio Flamingo: A Novel Audio Language Model with Few-Shot Learning
  and Dialogue Abilities}.
\newblock \emph{Proceedings of Machine Learning Research}, 235:25125--25148.

\bibitem[{Lippi and Torroni(2016)}]{Lippi2016}
Marco Lippi and Paolo Torroni. 2016.
\newblock \href {https://doi.org/10.1609/aaai.v30i1.10384} {{Argument mining
  from speech: Detecting claims in political debates}}.
\newblock \emph{30th AAAI Conference on Artificial Intelligence, AAAI 2016},
  pages 2979--2985.

\bibitem[{Liu et~al.(2025)Liu, Chen, Wang, Chen, Xu, Guo, Yang, Li, Yang, Jin,
  Fang, Zuo, Jionghao, and Liu}]{Liu2025}
Wenrui Liu, Qian Chen, Wen Wang, Yafeng Chen, Jin Xu, Zhifang Guo, Guanrou
  Yang, Weiqin Li, Xiaoda Yang, Tao Jin, Minghui Fang, Jialong Zuo, Bai
  Jionghao, and Zemin Liu. 2025.
\newblock \href {https://arxiv.org/abs/2505.24496} {Speech token prediction via
  compressed-to-fine language modeling for speech generation}.
\newblock \emph{Preprint}, arXiv:2505.24496.

\bibitem[{Mancini et~al.(2024{\natexlab{a}})Mancini, Ruggeri, Colamonaco,
  Zecca, Marro, and Torroni}]{Mancini2024a}
Eleonora Mancini, Federico Ruggeri, Stefano Colamonaco, Andrea Zecca, Samuele
  Marro, and Paolo Torroni. 2024{\natexlab{a}}.
\newblock \href {https://github.com/lt-nlp-lab-unibo/mamkit} {{MAMKit: A
  Comprehensive Multimodal Argument Mining Toolkit}}.
\newblock In \emph{Proceedings of the 11th Workshop on Argument Mining
  (ArgMining 2024)}, pages 69--82.

\bibitem[{Mancini et~al.(2022)Mancini, Ruggeri, Galassi, and
  Torroni}]{Mancini2022}
Eleonora Mancini, Federico Ruggeri, Andrea Galassi, and Paolo Torroni. 2022.
\newblock \href {https://aclanthology.org/2022.argmining-1.15/} {{Multimodal
  Argument Mining: A Case Study in Political Debates}}.
\newblock \emph{Proceedings of the 9th Workshop on Argument Mining}, pages
  158--170.

\bibitem[{Mancini et~al.(2024{\natexlab{b}})Mancini, Ruggeri, and
  Torroni}]{Mancini2024}
Eleonora Mancini, Federico Ruggeri, and Paolo Torroni. 2024{\natexlab{b}}.
\newblock {Multimodal Fallacy Classification in Political Debates}.
\newblock \emph{EACL 2024 - 18th Conference of the European Chapter of the
  Association for Computational Linguistics, Proceedings of the Conference},
  2:170--178.

\bibitem[{Mancini et~al.(2025)Mancini, Ruggeri, Villata, and
  Torroni}]{Mancini2025}
Eleonora Mancini, Federico Ruggeri, Serena Villata, and Paolo Torroni. 2025.
\newblock \href {https://doi.org/10.18653/v1/2025.argmining-1.35} {{Overview of
  MM-ArgFallacy2025 on Multimodal Argumentative Fallacy Detection and
  Classification in Political Debates}}.
\newblock In \emph{Proceedings of the 12th Argument mining Workshop}, pages
  358--368.

\bibitem[{Mestre et~al.(2023)Mestre, Middleton, Ryan, Gheasi, Norman, and
  Zhu}]{Mestre2023}
Rafael Mestre, Stuart~E. Middleton, Matt Ryan, Masood Gheasi, Timothy~J.
  Norman, and Jiatong Zhu. 2023.
\newblock \href {https://doi.org/10.18653/v1/2023.findings-eacl.21}
  {{Augmenting pre-trained language models with audio feature embedding for
  argumentation mining in political debates}}.
\newblock \emph{EACL 2023 - 17th Conference of the European Chapter of the
  Association for Computational Linguistics, Findings of EACL 2023}, (Section
  4):274--288.

\bibitem[{Mestre et~al.(2021)Mestre, Milicin, Middleton, Ryan, Zhu, and
  Norman}]{Mestre2021}
Rafael Mestre, Razvan Milicin, Stuart~E. Middleton, Matt Ryan, Jiatong Zhu, and
  Timothy~J. Norman. 2021.
\newblock \href {https://doi.org/10.18653/v1/2021.argmining-1.8} {{M-Arg:
  Multimodal Argument Mining Dataset for Political Debates with Audio and
  Transcripts}}.
\newblock \emph{8th Workshop on Argument Mining, ArgMining 2021 - Proceedings},
  (2014):78--88.

\bibitem[{Niizumi et~al.(2023)Niizumi, Takeuchi, Ohishi, Harada, and
  Kashino}]{Niizumi2023}
Daisuke Niizumi, Daiki Takeuchi, Yasunori Ohishi, Noboru Harada, and Kunio
  Kashino. 2023.
\newblock \href {https://doi.org/10.1109/TASLP.2022.3221007} {{BYOL for Audio:
  Exploring Pre-Trained General-Purpose Audio Representations}}.
\newblock \emph{IEEE/ACM Transactions on Audio Speech and Language Processing},
  31:137--151.

\bibitem[{Pittiglio(2025)}]{pittiglio-2025-leveraging}
Alessio Pittiglio. 2025.
\newblock \href {https://doi.org/10.18653/v1/2025.argmining-1.39} {{Leveraging
  Context for Multimodal Fallacy Classification in Political Debates}}.
\newblock In \emph{Proceedings of the 12th Argument mining Workshop}, pages
  388--397, Vienna, Austria. Association for Computational Linguistics.

\bibitem[{Radford et~al.(2021)Radford, Kim, Hallacy, Ramesh, Goh, Agarwal,
  Sastry, Askell, Mishkin, Clark, Krueger, and Sutskever}]{Radford2021}
Alec Radford, Jong~Wook Kim, Chris Hallacy, Aditya Ramesh, Gabriel Goh,
  Sandhini Agarwal, Girish Sastry, Amanda Askell, Pamela Mishkin, Jack Clark,
  Gretchen Krueger, and Ilya Sutskever. 2021.
\newblock \href {http://arxiv.org/abs/2103.00020} {{Learning Transferable
  Visual Models From Natural Language Supervision}}.

\bibitem[{Tahir et~al.(2025)Tahir, Ibrar, Ameer, Fatima, and
  Latif}]{tahir-etal-2025-prompt}
Abdullah Tahir, Imaan Ibrar, Huma Ameer, Mehwish Fatima, and Seemab Latif.
  2025.
\newblock \href {https://doi.org/10.18653/v1/2025.argmining-1.38}
  {{Prompt-Guided Augmentation and Multi-modal Fusion for Argumentative Fallacy
  Classification in Political Debates}}.
\newblock In \emph{Proceedings of the 12th Argument mining Workshop}, pages
  381--387, Vienna, Austria. Association for Computational Linguistics.

\bibitem[{Thimonier et~al.(2025)Thimonier, Perzo, and Seguier}]{Thimonier2025}
Hugo Thimonier, Antony Perzo, and Renaud Seguier. 2025.
\newblock \href {http://arxiv.org/abs/2508.14130} {{EmoSLLM:
  Parameter-Efficient Adaptation of LLMs for Speech Emotion Recognition}}.
\newblock pages 1--18.

\bibitem[{Vasuki and Vanathi(2006)}]{Vasuki2006}
A.~Vasuki and P.T. Vanathi. 2006.
\newblock \href {https://doi.org/10.1109/MP.2006.1664069} {A review of vector
  quantization techniques}.
\newblock \emph{IEEE Potentials}, 25(4):39--47.

\bibitem[{Vinciarelli et~al.(2009)Vinciarelli, Pantic, and
  Bourlard}]{Vinciarelli2009a}
Alessandro Vinciarelli, Maja Pantic, and Hervé Bourlard. 2009.
\newblock \href {https://doi.org/10.1016/j.imavis.2008.11.007} {{Social signal
  processing: Survey of an emerging domain}}.
\newblock \emph{Image and Vision Computing}, 27(12):1743--1759.

\bibitem[{Wolf et~al.(2019)Wolf, Debut, Sanh, Chaumond, Delangue, Moi, Cistac,
  Rault, Louf, Funtowicz, and Brew}]{Wolf2019}
Thomas Wolf, Lysandre Debut, Victor Sanh, Julien Chaumond, Clement Delangue,
  Anthony Moi, Pierric Cistac, Tim Rault, Rémi Louf, Morgan Funtowicz, and
  Jamie Brew. 2019.
\newblock \href {http://arxiv.org/abs/1910.03771} {{HuggingFace's Transformers:
  State-of-the-art Natural Language Processing}}.

\bibitem[{Zadeh et~al.(2018{\natexlab{a}})Zadeh, Liang, Poria, Cambria, and
  Morency}]{Zadeh2018b}
Amir Zadeh, Paul~Pu Liang, Soujanya Poria, Erik Cambria, and Louis-Philippe
  Morency. 2018{\natexlab{a}}.
\newblock \href {http://aclweb.org/anthology/P18-1208} {{Multimodal Language
  Analysis in the Wild: CMU-MOSEI Dataset and Interpretable Dynamic Fusion
  Graph}}.
\newblock \emph{Proceedings of ACL}, pages 2236--2246.

\bibitem[{Zadeh et~al.(2018{\natexlab{b}})Zadeh, Liang, Vanbriesen, Poria,
  Tong, Cambria, Chen, and Morency}]{Zadeh2018d}
Amir Zadeh, Paul~Pu Liang, Jonathan Vanbriesen, Soujanya Poria, Edmund Tong,
  Erik Cambria, Minghai Chen, and Louis~Philippe Morency. 2018{\natexlab{b}}.
\newblock \href {https://doi.org/10.18653/v1/p18-1208} {{Multimodal language
  analysis in the wild: CMU-MOSEI dataset and interpretable dynamic fusion
  graph}}.
\newblock \emph{ACL 2018 - 56th Annual Meeting of the Association for
  Computational Linguistics, Proceedings of the Conference (Long Papers)},
  1:2236--2246.

\bibitem[{Zhang et~al.(2024)Zhang, Zhang, Li, Zhou, and Qiu}]{Zhang2024}
Xin Zhang, Dong Zhang, Shimin Li, Yaqian Zhou, and Xipeng Qiu. 2024.
\newblock {SPEECHTOKENIZER: UNIFIED SPEECH TOKENIZER FOR SPEECH LANGUAGE
  MODELS}.
\newblock In \emph{12th International Conference on Learning Representations,
  ICLR 2024}, pages 1--21.

\end{thebibliography}

\appendix

\section{Dataset Details} \label{app:datasets}

The \textbf{MM-USED-Fallacy} dataset \cite{Mancini2024,Mancini2025} contains 18,910 utterances from official US presidential elections (USElecDeb60to16; \citealt{Haddadan2019}), manually annotated for fallacy presence, with 1,457 fallacious utterances. Audio utterances have an average duration of 10.4s.

For AFC, fallacies were annotated as argumentative fallacies in six categories \cite{Goffredo2022}: Ad Hominem, Appeal to Authority, Appeal to Emotion, False Cause, Slippery Slope, and Slogan. 
The class distribution is unbalanced in the official test sets: Slogan (4\%), Appeal to Authority and False Cause (7\% each), and no Slippery Slope examples. In AFD, the fallacy ratio is 11.77\%.

The \textbf{CMU-MOSEI} dataset \cite{Zadeh2018d} contains more than 23,500 annotated utterances from over 1,000 online YouTube videos. The sentiment analysis task is a regression problem with scores ranging from -3 to +3, reformulated as classification following the authors' protocol. The emotion recognition task is multi-label classification with six basic emotions (happy, sad, angry, fear, disgust, surprise).

\section{Implementation details} \label{app:impl}

The \texttt{transformers} library \cite{Wolf2019} was used to access the pre-trained \texttt{Llama-3.2-3B}, \texttt{Qwen2-Audio-7B-Instructed}, and \texttt{Qwen2-Audio-7B} as well as to fine-tune the models. 
%
Experiments were run using torch 2.1.2 \cite{Abadi2016}, transformers 4.46.3 \cite{Wolf2019}, a GPU Nvidia RTX-A6000 and CUDA 12.2. 
Baselines come from existing papers \cite{Mancini2025,pittiglio-2025-leveraging,tahir-etal-2025-prompt,cantin-larumbe-chust-vendrell-2025-argumentative}, or were computed for this work. The ICL using Qwen2-Audio has been computed in a 3-shot way. The prompts are available in Appendix \ref{app:prompts_q2a}. 
The BYOL-A v2 audio representations have been generated using the pre-trained model, following the protocol of \citet{Niizumi2023}, taking 10s of audio by padding or random cropping. The projection to the LLM embedding space has been done with a fully connected layer.

\section{In-Context-Learning prompts}
\label{app:prompts_q2a}

The prompts used for the 3-shot In-Context-Learning classification with Qwen2-Audio is shown in Figure \ref{fig:prompt_q2a}.

\begin{figure*}[ht]

\centering
\begin{tcolorbox}[colback=orange!5!white, colframe=black, rounded corners, width=0.95\linewidth]
\textit{Prompt}
\small 

\vspace{0.2em}
\texttt{You are a highly capable assistant specialized in audio processing tasks. You receive inputs containing audio token representations followed by text instructions, and return structured answer.} \\

\texttt{You may be asked to perform: }

\texttt{**Automatic Fallacy Classification (AFC)** — identify the type of fallacy expressed in the audio.} \\

\texttt{Follow this output formats:}

\texttt{- For AFC task: ’| FallacyClass: <Fallacy code> |’}\\

\texttt{For AFC task, Fallacy must be provided as two-letter codes chosen from the following fallacy codes. Here they are with the definition of fallacies: } \\

\texttt{- **AA** (Appeal to Authority):  Refers to the use of an expert’s opinion as evidence to back up an argument.}

\texttt{- **AH** (Ad Hominem): Is characterized by an excessive attack on an arguer’s position. }

\texttt{- **AE** (Appeal to Emotion):  Usually involves the use of loaded language. }

\texttt{- **FC** (False Cause):  Regards the misinterpretation of correlation as causation. }

\texttt{- **SL** (Slogan): Is brief and striking phrase used to evoke excitement. }

\texttt{- **SS** (Slippery Slope): Is an argument that claims exaggerated outcomes for a given action. } \\

\texttt{Perform the following audio-based tasks:
**Automatic Fallacy Classification (AFC)**.}

\texttt{Classify the Fallacy of the text and audio into one of the different classes you were given.} \\

\texttt{Remember: } 

\texttt{1. You ought to answer ONLY using the two-letter codes and stop your generation. }

\texttt{2. Only use the classes that you were given, not other ones.}

\texttt{3. Always output one of the following: AA, AH, AE, FC, SL or SS.} \\

\texttt{Some examples are:}

\texttt{- "We are the most admired country in the world.": AE}

\texttt{- "And then long-term we've got to fix our health care system, we've got to fix our energy system that is putting such an enormous burden on families.": AE}

\texttt{- "You folks home will have an empty chair at the kitchen table this morning. That man or wife going to bed tonight and reaching over to try to touch their, out of habit, where their wife or husband was, is gone.": AE}

\texttt{- "And Colin Powell so stated.": AA}

\texttt{- "I think it was one of the things that Mr. Bush said about Mr. Reagan back in 1980.": AA}

\texttt{- "Well, here's the problem with it: It sounds very good, but there's a reason that 500 economists, including seven Nobel prize winners and business periodicals like Business Week, and even Senator Dole's friends, Senator Warren Rudman, former Republican senator from New Hampshire, says it's not a practical program.": AA}

\texttt{- "Bill Clinton has trouble telling the truth.": AH}

\texttt{- "First, maybe he's not as rich as he says he is.": AH}

\texttt{- "This guy's three years younger and a lot less competent.": AH}

\texttt{- "He's not going to play games with me.": FC}

\texttt{- "We find that the prices you pay went up five times as much in the Truman Administration as they did in the Eisenhower Administration.": FC}

\texttt{- "We've got two and a half million more Americans out of work now than we had when Mr. Ford took office.": FC}

\texttt{- "They'd impose the Green New Deal, which would crush American energy, would increase the energy cost of American families in their homes and literally would crush American jobs.": SS}

\texttt{- "Although Governor Reagan has changed his position lately, on four different occasions, he has advocated making Social Security a voluntary system, which would, in effect, very quickly bankrupt it.": SS}

\texttt{- "And if the agricultural economy collapses, then the economy of the rest of the United States sooner or later will collapse.": SS}

\texttt{- "It's time to turn the page.": SL}

\texttt{- "Our character is on the ballot.": SL}

\texttt{- "The American people are moving again, and moving in the right direction.": SL}

\end{tcolorbox}
\caption{Task prompt for the Qwen2-Audio model for In-Context-Learning}
\label{fig:prompt_q2a}
\end{figure*}

\section{Manual Analysis of Examples}

\paragraph{AFC task} We found out that the multimodal model was overall making less confusion between the appeal to authority and appeal to emotion classes. When it detected better appeal to authority, the voice was assertive and monotonous, whereas for appeal to emotion the voice was more prosodic, for example with tremor (i.e., higher Jitter) in the beginning of the utterance. We believe this is due to the fact that an emotional voice would have a specific prosody and other cues related to non semantic speech. 

\paragraph{EMO task} The findings were similar. Examples where text is neutral are difficult to predict using the unimodal model. Other example where the information is contextual and the acoustic modality gives the sufficient context: "\textit{If you wanna see a good superhero, if you wanna see a good superhero movie, go see any of the Spider-Mans, go see any of the go see Batman Begins, that's a good one[...]}" was misclassified as happy with text but well classified as disgust by the multimodal model. 

\section{Baseline Descriptions} \label{app:baselines}

\paragraph{Text-only (Llama3-3B SFT)}
The LLaMA-3.2-3B model fine-tuned with LoRA on text transcripts only, using the same training protocol as our full method (Section~\ref{subsec:downstream}).

\paragraph{Text + Audio (Llama3-3B + BYOL-A)}
The text-only model augmented with a BYOL-A v2 \cite{Niizumi2023} audio representation---a single 2048-dimensional vector per utterance that temporally aggregates the audio signal. This vector is projected onto the LLM embedding space via a learned linear layer and concatenated with the sequence of text token embeddings.

\paragraph{Qwen2-Audio-7B}
A 7B-parameter multimodal model \cite{Chu2024} that integrates audio via Whisper encoder representations. We evaluate it in both In-Context-Learning (ICL, 3-shot) and Supervised Fine-Tuning (SFT) settings.

\paragraph{Shared-task baselines ($^\dagger$)}
Results marked with $^\dagger$ are taken directly from \citet{Mancini2025} and use lightweight encoder architectures (RoBERTa + various audio encoders) with task-specific fusion layers.

\paragraph{Using all audio tokens}
We note that using \emph{all} audio tokens without selection is computationally impractical ($\sim$4{,}000 tokens for 10\,s of speech) and did not yield positive results in our preliminary experiments, consistent with findings in \citet{Mancini2024a}.

\end{document}